\documentclass[runningheads]{llncs}
\usepackage{graphicx}
\usepackage{subcaption} 

\usepackage{booktabs}

\usepackage[ruled,vlined,linesnumbered]{algorithm2e}

\usepackage{amssymb}
\usepackage{amsmath}

\usepackage{orcidlink}

\newcommand*\samethanks[1][\value{footnote}]{\footnotemark[#1]}

\begin{document}
\title{Explainable AI for Fair Sepsis Mortality Predictive Model\thanks{This manuscript has been accepted to the 22nd International Conference on Artificial Intelligence in Medicine (AIME'24).}}
%
%
\author{Chia-Hsuan Chang\thanks{Both authors contributed equally.}\orcidlink{0000-0001-9116-8244} \and
Xiaoyang Wang\samethanks\orcidlink{0000-0002-8471-4670} \and
Christopher C. Yang\orcidlink{0000-0001-5463-6926}}
\authorrunning{Chang et al.}
%
\institute{College of Computing \& Informatics, Drexel University, Philadelphia PA 19104, USA\\
\email{\{cc3859,xw388,chris.yang\}@drexel.edu}}
\maketitle              
\begin{abstract}
Artificial intelligence supports healthcare professionals with predictive modeling, greatly transforming clinical decision-making. This study addresses the crucial need for fairness and explainability in AI applications within healthcare to ensure equitable outcomes across diverse patient demographics. By focusing on the predictive modeling of sepsis-related mortality, we propose a method that learns a performance-optimized predictive model and then employs the transfer learning process to produce a model with better fairness. Our method also introduces a novel permutation-based feature importance algorithm aiming at elucidating the contribution of each feature in enhancing fairness on predictions. Unlike existing explainability methods concentrating on explaining feature contribution to predictive performance, our proposed method uniquely bridges the gap in understanding how each feature contributes to fairness. This advancement is pivotal, given sepsis's significant mortality rate and its role in one-third of hospital deaths. Our method not only aids in identifying and mitigating biases within the predictive model but also fosters trust among healthcare stakeholders by improving the transparency and fairness of model predictions, thereby contributing to more equitable and trustworthy healthcare delivery. 

\end{abstract}
\section{Introduction}

Artificial intelligence (AI) has the potential to revolutionize clinical decision-making by providing support to clinicians. Predictive modeling, a common AI technique, automates and accelerates various healthcare tasks, such as readmission, mortality, and disease recognition. However, it is crucial to prioritize fairness in these algorithms to ensure that the benefits of AI decision-making are distributed equitably across diverse patient populations. This helps to minimize the risk of perpetuating biases and inequalities in healthcare outcomes. Additionally, explainable AI (XAI) is one of the promising research topics in healthcare informatics~\cite{combiIHIRochesterReport2023}. XAI explains the model's prediction, assisting clinicians and practitioners in gaining insights from data rather than mindlessly following the prediction outcomes~\cite{yangExplainableArtificialIntelligence2022}. Enhancing fairness and explainability gains more trust from the users \cite{angerschmidEffectsFairnessExplanation2022,zhouExplainabilityAIFairness2022} and is a key element of trustworthy AI designed for the healthcare system~\cite{combiIHIRochesterReport2023}.

While various XAI techniques, such as SHapley Additive exPlanations (SHAP)~\cite{lundbergUnifiedApproachInterpreting2017}, are available to explain the contribution of features to predictive performance, there remains a gap in explaining feature contributions to fairness enhancement. Understanding how specific features impact fairness is crucial, as this can help identify areas where biases may exist within the model and enable targeted interventions to mitigate these disparities. Therefore, exploring methods that provide explainability and facilitate the evaluation on fairness enhancement within clinical predictive models is essential.

Sepsis is one of the foremost life-threatening medical emergencies encountered by critically ill patients in the United States, arising as a complex syndrome that epitomizes the intricate interplay among the infecting microorganism, and the host's immune and coagulation responses. Annually, sepsis afflicts over 1.7 million adults in the United States, leading to a stark mortality rate wherein more than 350,000 of these individuals succumb during their hospital stay or are transitioned to hospice care~\cite{cdcSepsisBodyExtreme2023}. Remarkably, sepsis is implicated in one out of every three hospital fatalities, underscoring its critical impact on healthcare outcomes. Given the pressing need to enhance clinical decision-making in the management of sepsis—a condition with profound implications for patient mortality—this study proposes a method, which integrates the principles of fairness and explainability into machine learning models tailored for sepsis mortality predictions. 

The contributions of our proposed method are twofold: (1) we learn a fair sepsis mortality predictive model for different races from a performance-optimized predictive model by applying transfer learning, and (2) we introduce a novel permutation-based feature importance algorithm. Our method not only enables a nuanced understanding of how each individual feature contributes to enhancing fairness across different races but also provides healthcare practitioners with new insights and assistance. As far as we know, this is the first attempt to explain how features contribute to improving fairness. In our research, we extend the conventional focus on bias mitigation within predictive models to encompass a detailed examination of the relationship between features and fairness. By doing so, we bridge the realms of model explainability and fairness. Our findings contribute to the improvement of clinical outcomes, the reinforcement of ethical standards in healthcare AI, and the promotion of trust and transparency among medical professionals and patients alike.

\section{Related Work} 

The process of explaining the rationale behind predictive modeling is crucial in ensuring fairness in human judgment~\cite{zhouExplainabilityAIFairness2022}. AI-generated explanations can help identify factors that contribute to unfair outcomes. Both fairness and explanation are crucial aspects of creating trustworthy and responsible AI, and research shows that they are positively associated with user trust in scenarios where AI is used to make decisions~\cite{angerschmidEffectsFairnessExplanation2022}. However, the current approaches, LIME~\cite{ribeiroWhyShouldTrust2016} and SHAP~\cite{lundbergUnifiedApproachInterpreting2017}, only explain the contribution of features to predictive performance. As far as we know, no study yet proposes a method for explaining the contribution of features to fairness.

Currently, the researchers have devoted considerable efforts toward developing predictive models for sepsis mortality. Bao et al.~\cite{baoMachinelearningModelsPrediction2023} and Taylor et al.~\cite{taylorPredictionInhospitalMortality2016} explored a diverse array of models and features with the aim of identifying the model that exhibits optimal performance in predicting sepsis-related mortality and implemented feature importance analysis that elucidates the influence of individual features on the overall prediction performance. Wang et al.~\cite{wangComparisonMachineLearning2022} have elucidated that disparities rooted in social determinants among groups of sepsis patients by various currently available diagnostic criteria and the efficacy of risk prediction models for sepsis is notably diminished when a universally trained model is indiscriminately applied across different subpopulations. However, these endeavors have predominantly aimed at enhancing prediction accuracy and have not conducted a thorough and precise analysis of explainability and fairness, leading to a situation where the predictive results, though accurate, may lack transparency and equitable consideration across diverse patient demographics.

\section{Materials and Methods}

\subsection{Dataset Preparation}

Our study focused on the critical task of predicting mortality among sepsis patients. The dataset we used is extracted from the MIMIC-IV database, which encompasses comprehensive medical records and demographic information of individuals diagnosed with sepsis. The Medical Information Mart for Intensive Care IV(MIMIC-IV) dataset~\cite{johnson_mimic-iv_nodate} is a comprehensive collection of critical care records, detailing patient admissions to the Intensive Care Units (ICU) at Beth Israel Deaconess Medical Center. We preprocessed the dataset to make the experimental result reasonable and robust. In this study we measured the fairness by considering the predictive performance difference between races, so we removed the samples without a definitive race record and transformed the categorical race variable into a binary format: White vs Non-White. To avoid potential timely treatment bias, for multiple-time medical measurement records, we used the average of the first 24-hour records after the first ICU admission. To eliminate the extreme cases, the patient record exhibiting a value for at least one feature that fell outside the interpercentile range of [2\%, 98\%] for the total population was removed. Therefore, the remaining features are Age, Gender, Race, Temperature, Weight, Heart Rate, Glucose, Systolic Blood Pressure(SBP), Diastolic Blood Pressure(DBP), SpO2, Respiratory rate, Renal Replacement Therapy (RRT), Sequential Organ Failure Assessment (SOFA),  Charlson Comorbidity Index (CCI), and APACHE-III score. The target label is the death of the patient, positive means the patient died, otherwise the patient survived. Table \ref{tab. distribution} presents the descriptive statistics of the dataset employed in our study, offering the detailed distribution of target variables. 

\begin{table}
\centering
\caption{The MIMIC-IV Dataset Distribution}
\label{tab. distribution}
\begin{tabular}{llll}
\toprule
Characteristic~ & \begin{tabular}[c]{@{}l@{}}Overall\\ (N = 9,349)\end{tabular} & \begin{tabular}[c]{@{}l@{}}Neg. Class\\ (N = 7,806)\end{tabular} & \begin{tabular}[c]{@{}l@{}}Pos. Class\\ (N = 1,543)\end{tabular} \\ 
\midrule  
Race\\   \hspace{1ex}White & 7,797 (83.4\%)~  & 6,546 (83.9\%)~ & 1,251 (81.1\%)~\\   \hspace{1ex}Non-White & 1,552 (16.6\%) &  1,260 (16.1\%) & ~~292 (18.9\%) \\
\bottomrule
\end{tabular}
\end{table}

\subsection{Predictive Model and Fair Model}

\begin{figure}
    \centering
    \includegraphics[width=0.85\textwidth]{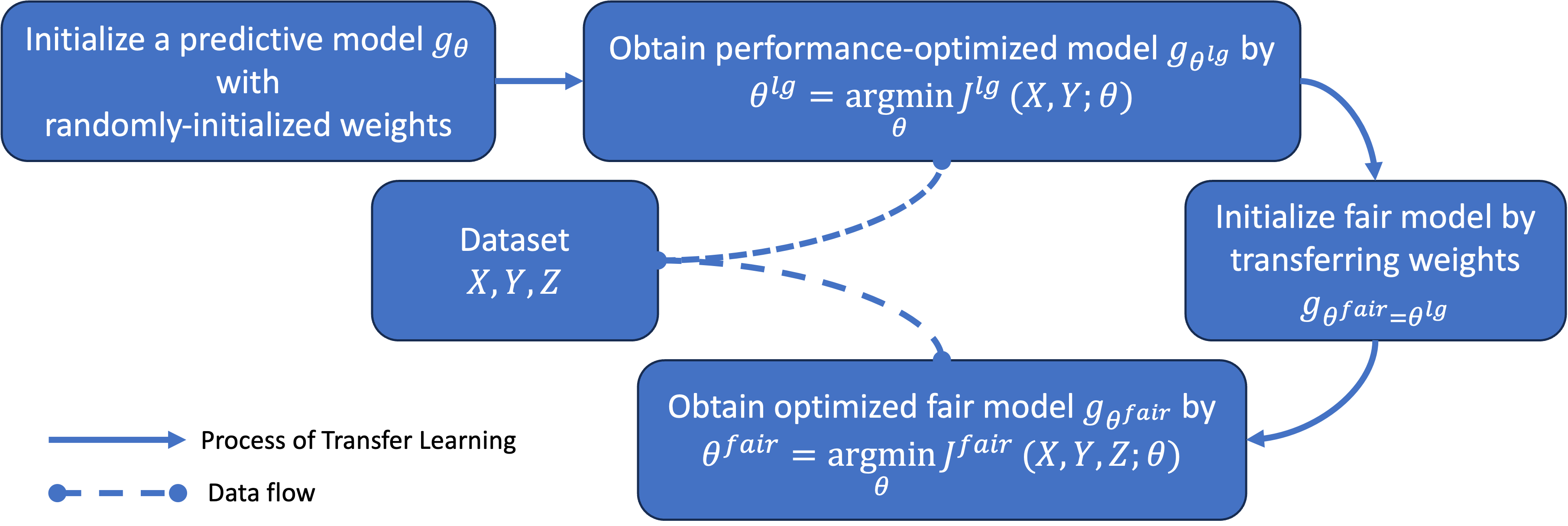}
    \caption{The proposed transfer learning process for performance-optimized and fair model}
    \label{fig:transferring-process}
\end{figure}

Fig.~\ref{fig:transferring-process} illustrates the process to obtain performance-optimized predictive model and fair model, detailed as follows.

Given the dataset~$X \in \mathbb{R}^{n \times m}$, we train a logistic regression model as the mortality predictive model~$g_{\theta^{lg}}$, where $n$ is the number of septic patients and $m$ is the number of features. We optimize feature weights $\theta^{lg} \in \mathbb{R}^m$ to minimize the binary cross-entropy between ground truths and predictions:

\begin{equation} \label{eq.lg_loss}
    J^{lg}(X, Y; \theta) = - \sum^{n}_{i=1} [Y_ilog(g_{\theta}(X_i)) +(1-Y_i)log(1-g_{\theta}(X_i))],
\end{equation}

\noindent where ground truth $Y \in \mathbb{R}^n$ is al a list of dichotomous values, and $Y_i$ indicates whether the patient $i$ is dead (1) or not (0), and $g_{\theta}(X_i)$ is a outcome prediction to patient $i$. By minimizing Eq.~\ref{eq.lg_loss}, the resultant model, $g_{\theta^{lg}}$, is only performance-optimized for predictive performance. To make it generate the fair prediction for different races, we implement a transfer learning process that initializes a fair model by transferring weights $\theta^{lg}$ as fair model's initial weights~$g_{\theta^{fair} = \theta^{lg}}$. The idea behind the weight transferring is that fair model inherits the pre-learned capabilities, ensuring its predictive ability is close to $g_{\theta^{lg}}$. Furthermore, we design a different optimization goal for fair model to make its predictions become fair across races. Specifically, we optimize the fair model for minimizing the Equalized Odds Disparity (EOD) between races. EOD is a comprehensive fairness measurement~\cite{hardtEqualityOpportunitySupervised2016}, denoted $\mathcal{E}$,  that sums the square difference in true positive rate (TPR) and false positive rate (FPR) between races:

\begin{align} \label{eq.eod}
    \mathcal{E}(g_{\theta}(X), Y, Z) &= TPR_{Diff} + FPR_{Diff} \\
    TPR_{Diff} &= (Pr(g_{\theta}(X)=1 | Y=1, Z=1) -  Pr(g_{\theta}(X)=1 | Y=1, Z=0))^2 \nonumber \\
    FPR_{Diff} &= (Pr(g_{\theta}(X)=1 | Y=0, Z=1) -  Pr(g_{\theta}(X)=1 | Y=0, Z=0))^2 \nonumber
\end{align}

\noindent where $Z=1$ for White and $Z=0$ for Non-White, respectively. To prevent losing the predictive performance, we add a regularization term $\Omega$, allowing overall TPR performance reached by performance-optimized predictive model $g_{\theta^{lg}}$ be changed in a small range: $\Omega(X, Y; \theta) = TPR - \epsilon < TPR < TPR + \epsilon$, where $TPR=Pr(g_{\theta}(X) = 1 | Y=1)$. As a result, the loss function turns out to be:

\begin{equation} \label{eq.eod_loss}
    J^{fair}(X, Y, Z; \theta) = \mathcal{E}(g_{\theta}(X), Y, Z) + \Omega(X, Y; \theta),
\end{equation}

\noindent we treat the optimized weights $\theta^{fair}$ as the final weights for the fair model, $g_{\theta^{fair}}$. 

\subsection{Permutation Fairness Importance Algorithm}

Permutation-based feature importance method~\cite{breimanRandomForests2001} was developed to quantify how each feature contributes to the predictive performance, but it does not measure how the features contribute to fairness when a fair model is compared with a base model. 

To quantify the influence of each feature on the fairness improvement, we first define the fairness improvement, which calculates the difference of EOD performance ($\mathcal{E}$, Eq.~\ref{eq.eod}) between $g_{\theta^{lg}}$ and $g_{\theta^{fair}}$:

\begin{equation}
    \mathcal{F}(X, Y, Z, g_{\theta^{lg}}, g_{\theta^{fair}}) = \mathcal{E}(g_{\theta^{fair}}(X), Y, Z) - \mathcal{E}(g_{\theta^{lg}}(X), Y, Z)
\end{equation}

\noindent Note that the smaller EOD, the smaller the difference in predictive performance between different races. In our experimental results, we observe that fair model $g_{\theta^{fair}}$ always has a smaller EOD than performance-optimized model $g_{\theta^{lg}}$. Therefore, the more negative $\mathcal{F}(\cdot)$ suggests more fairness improvements achieved by the fair model $g_{\theta^{fair}}$. 

To quantify the contribution of each feature~$i$ on the fairness improvement, we extend the permutation-based feature importance algorithm and propose the Algorithm~\ref{alg:permutation-fairness-importance-algorithm}. Specifically, we define a $permute$ function which randomly shuffles the values for the feature $i$ (i.e., column $i$ in $X$) and keeps the values of the other features unchanged. Since the $permute$ function involves randomness, we reiterate it on the feature $i$ for $J$ times, resulting in $J$ different permuted $\tilde{X}^j, j \in [1,...,J]$ (line $3-4$). We then calculate the influence of the feature $i$ on fairness improvement, denoted $\delta_i$, by averaging $\mathcal{F}(\tilde{X}^j, Y, Z, g_{\theta^{lg}}, g_{\theta^{fair}})$ over $J$ permuted inputs (line~$5$). The larger magnitude of $\delta_i$ suggests that feature~$i$ contributes more to the fairness improvement, also demonstrating its importance on fairness.

\begin{algorithm}
\caption{Permutation Fairness Importance Algorithm}\label{alg:permutation-fairness-importance-algorithm}
\SetKwInOut{Input}{Require}
\SetKwInOut{Output}{Return}
\Input{Dataset $X$, ground truths $Y$, race indicators $Z$, performance-optimized predictive model $g_{\theta^{lg}}$, and fair predictive model $g_{\theta^{fair}}$}

initialize empty list of importance score~$\delta$

\For{each feature $i$ (column of $X$)}{
    \For{each repetition $j \in 1,...,J$}{
      $\tilde{X}^j = permute(X[:,i])$
    }
    $\delta_i = \frac{1}{J}\sum^{J}_{j=1}\mathcal{F}(\tilde{X}^j, Y, Z, g_{\theta^{lg}}, g_{\theta^{fair}})$
}
\Return $\delta$
\end{algorithm}

\subsection{Evaluation Setup}

We employed the 5-fold cross-validation approach, and in each fold we conducted feature permutation 100 ($J=100$) times to ensure the robustness of our findings. For the evaluation of model performance and fairness, we utilized a comprehensive suite of metrics including the Area Under the Receiver Operating Characteristic curve (AUC), Accuracy (ACC), group differences in TPR (TPR diff) and FPR (FPR diff), and the EOD. To choose the classification threshold for $g_{\theta^{lg}}$ and $g_{\theta^{fair}}$, we adopted the ``The Closest to (0,1) Criteria (ER)", which selects the threshold that minimizes the Euclidean distance between ROC curve and the ideal point of (0,1) on the ROC space~\cite{unalDefiningOptimalCutPoint2017}. Based on these settings, our result analysis comprises several pivotal elements: Firstly, the feature importance is assessed using SHAP and we analyze the coefficient changes between $\theta^{lg}$ and $\theta^{fair}$, which provides the detailed changes of each feature from the performance-optimized model to fair model. Furthermore, our analysis extends to examining the differential fluctuations in EOD across different races with feature permutation, identifying the extent to which specific features contribute to the model fairness. 

\section{Results and Discussions}

\textbf{Fair model has better fairness performance than performance-optimized predictive model while having a decreased predictive performance.} Table~\ref{tab:performance-fairness-table} reports predictive and fairness performance comparison between performance-optimized predictive model $g_{\theta^{lg}}$ and fair model $g_{\theta^{fair}}$. The fair model has smaller TPR and FPR differences across demographic groups, leading to a smaller EOD compared with the performance-optimized predictive model. We observe that for fair model both predictive performances, including AUC and Accuracy, drop for exchange the fairness improvement. 

\begin{table}[h]
\centering
\caption{Predictive and fairness performance comparison}
\label{tab:performance-fairness-table}
\resizebox{0.65\textwidth}{!}{%
\begin{tabular}{@{}lccccc@{}}
\toprule
 & AUC & ACC & $TPR_{Diff}$ & $FPR_{Diff}$ & EOD ($\mathcal{E}$) \\ \midrule
$g_{\theta^{lg}}(X)$ & 0.746 & 0.675 & 0.071 & 0.064 & 0.068 \\
$g_{\theta^{fair}}(X)$ & 0.723 & 0.637 & 0.002 & 0.015 & 0.008 \\ \bottomrule
\end{tabular}%
}
\end{table}

\begin{figure}[h]
    \centering
    \begin{subfigure}[b]{0.495\linewidth}
        \includegraphics[width=\linewidth]{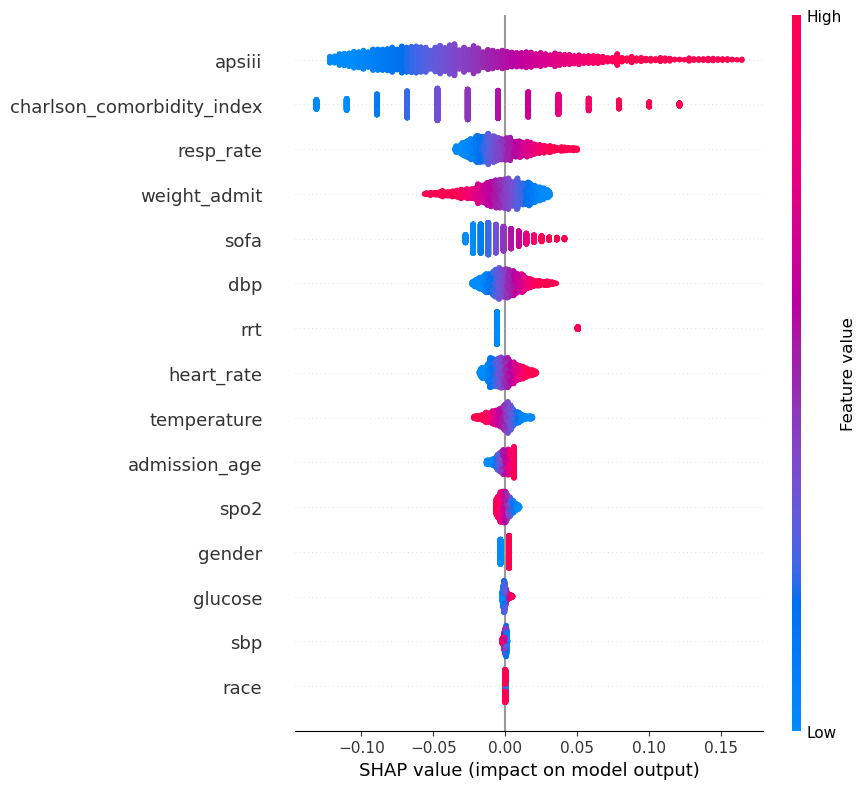}
        \label{fig:plot1}
    \end{subfigure}
    \begin{subfigure}[b]{0.495\linewidth}
        \includegraphics[width=\linewidth]{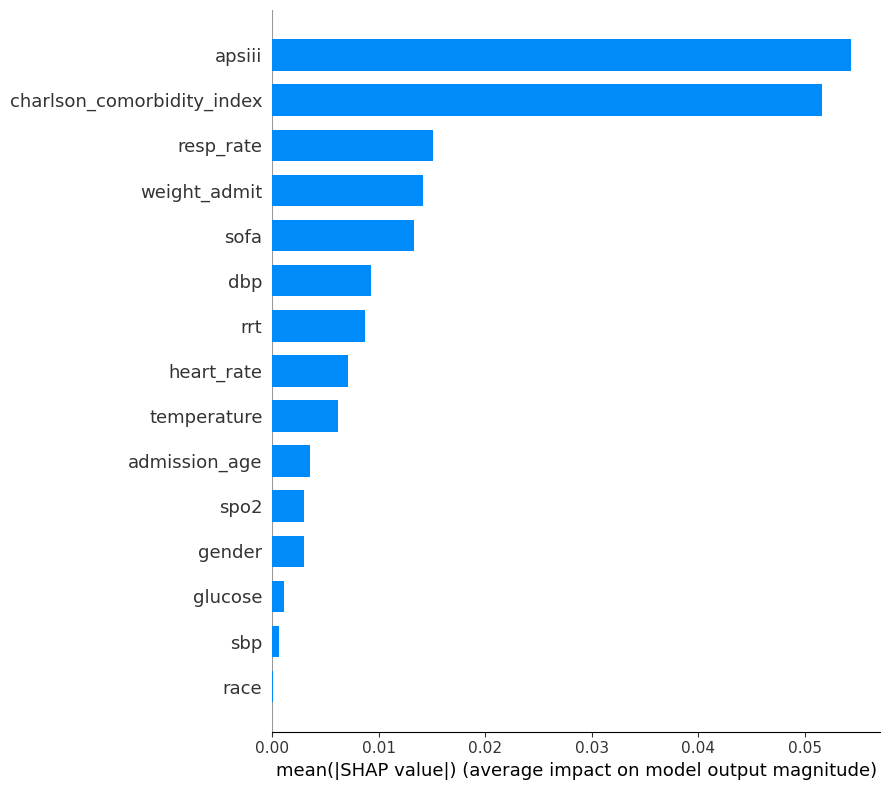}
        \label{fig:plot2}
    \end{subfigure}
    \begin{subfigure}[b]{0.495\linewidth}
        \includegraphics[width=\linewidth]{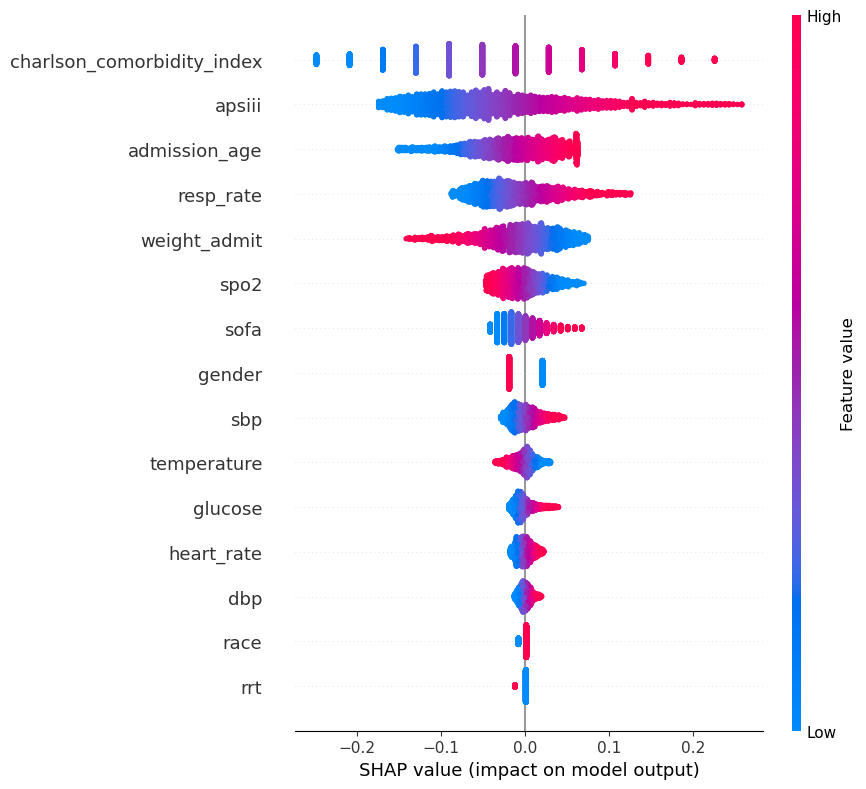}
        \label{fig:plot3}
    \end{subfigure}
    \begin{subfigure}[b]{0.495\linewidth}
        \includegraphics[width=\linewidth]{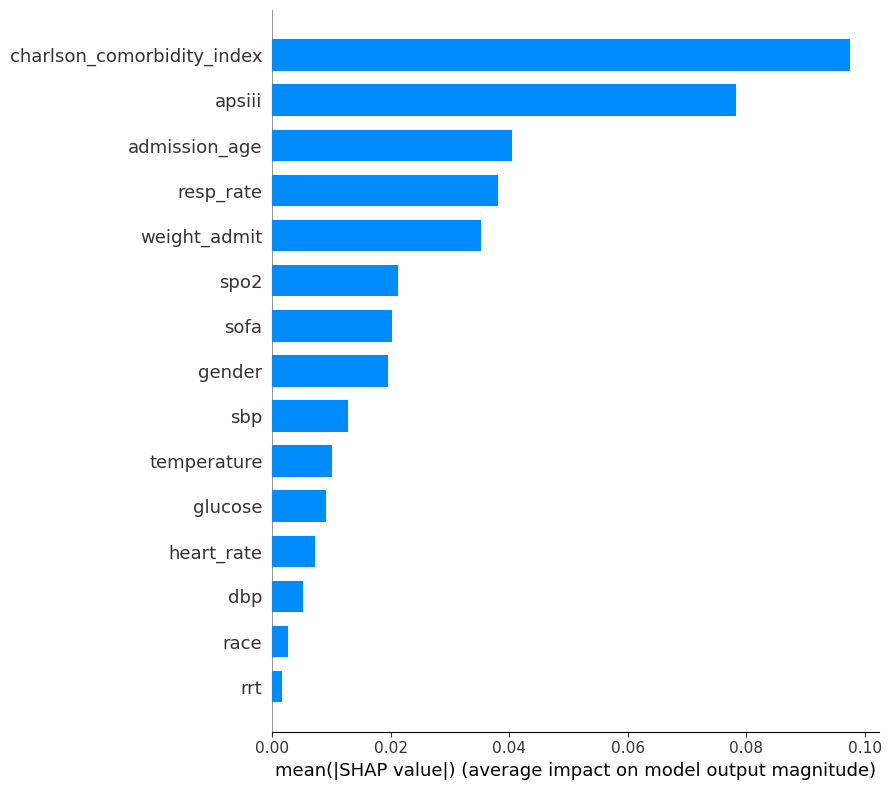}
        \label{fig:plot4}
    \end{subfigure}
    \caption{The SHAP values and importance for performance-optimized predictive model (two upper subplots) and fair model (two lower subplots).}
    \label{fig:shap-rankings}
\end{figure}

\textbf{What are the important predictive variables for performance-optimized predictive model and fair model?} Fig.~\ref{fig:shap-rankings} reports SHAP values and feature importance rankings for performance-optimized predictive model (upper two subplots) and fair model (lower two subplots). We notice that two most important variables, including \textit{aspiii} and \textit{charlson comorbidity index}, remain the same between two models while the importance of few variables change significantly. For example: (1) \textit{age} and \textit{spo2} contribute more in fair model than in performance-optimized predictive model, and (2) fair model depends more on \textit{sbp} than \textit{dbp}; on the contrary, performance-optimized predictive model relies more on \textit{dbp} than \textit{sbp} in prediction. These observations suggest that when optimizing performance-optimized predictive model for better fairness performance, the model will alter its reliance on the features, resulting in changes of feature importance.  

\begin{figure}[h]
    \centering
    \begin{subfigure}[b]{0.495\linewidth}
        \includegraphics[width=\linewidth]{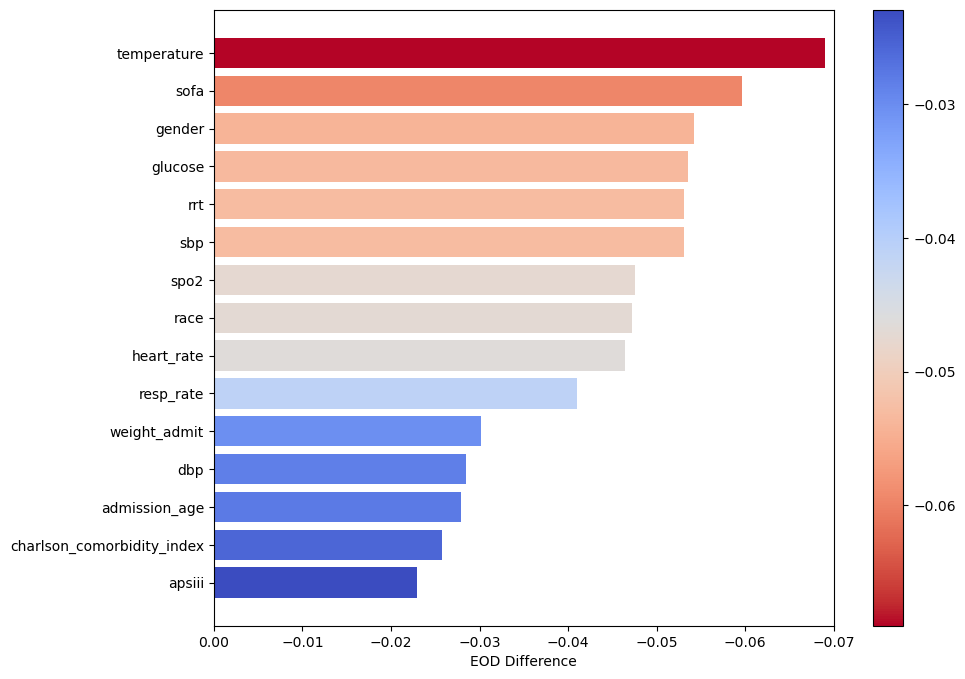}
        \caption{Features vs EOD Differences}
        \label{fig:feature_vs_EOD_Diff}
    \end{subfigure}
    \begin{subfigure}[b]{0.495\linewidth}
        \includegraphics[width=\linewidth]{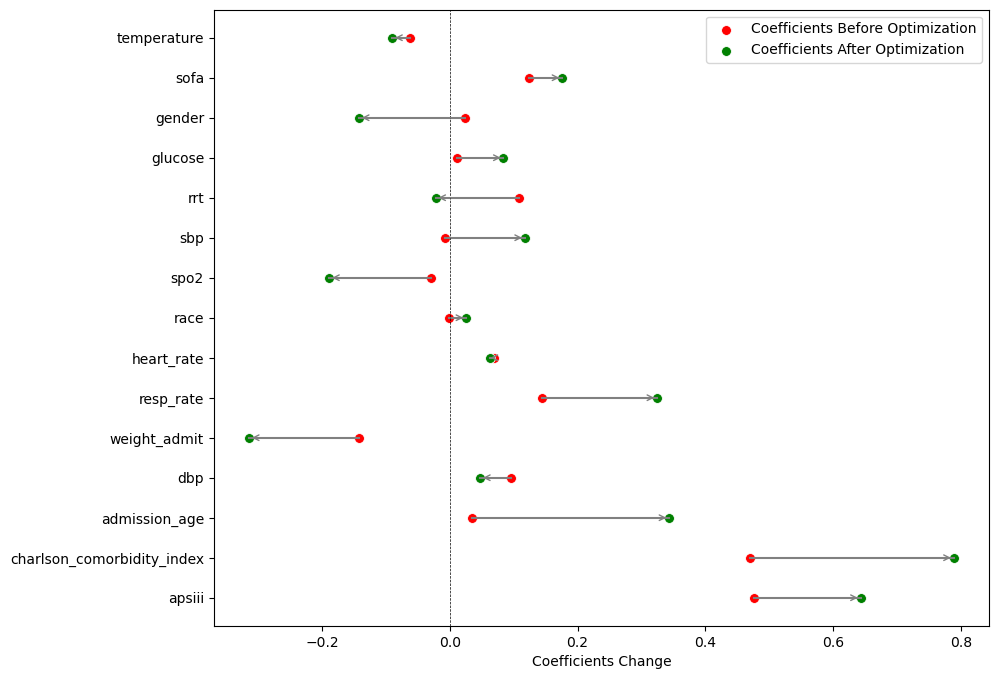}
        \caption{Coefficients Changes}
        \label{fig:coef_changes}
    \end{subfigure}
    \caption{Comparison of fairness contribution (left subplot) and coefficients change (right subplot) of each feature.}
\end{figure}

\textbf{What are the important features that influence the fairness?} SHAP values only tell us how each feature of the model contributes to predictive performance. Based on our proposed algorithm~(Algorithm~\ref{alg:permutation-fairness-importance-algorithm}), Fig.~\ref{fig:feature_vs_EOD_Diff} shows how each feature contributes to the EOD difference. The feature of the more negative EOD difference suggests it has a higher influence on reducing the prediction disparity between races (Whites and Non-Whites). We observe that there are four sets of features grouped by the value gaps. For features having EOD Difference $< -0.06$, the first feature set has the two most influential variables on fairness performance. They are \textit{temperature} (rank 10) and \textit{sofa} (rank 7) with an average rank of 8.50, where rank is referred their SHAP importance rankings in the lower-right subplot of Fig.~\ref{fig:shap-rankings}. For those having EOD Difference between $[-0.05, -0.06]$, the second set has \textit{gender} (rank 8), \textit{glucose} (rank 11), \textit{rrt} (rank 15), and \textit{sbp} (rank 9) with average rank 10.75. The third feature set, EOD Difference falls between $[-0.04, -0.05]$, consists of \textit{spo2} (rank 6), \textit{race} (rank 13), and \textit{heart\_rate} (rank 11), and \textit{resp\_rate} (rank 4) with average rank 8.50. The last set contains the least influential variables for fairness performance because their EOD differences are $> -0.03$. They are \textit{weight\_admit} (rank 5), \textit{dbp} (rank 13), \textit{age} (rank 3), \textit{charlson comorbidity index} (rank 1), and \textit{aspiii} (rank 2) with average rank 4.80. Our rank analysis suggests that the most influential variables for the fairness performance (i.e., the first feature set) are not the variables that highly contribute to the predictive performance; on the contrary, those least influential variables for fairness performance (i.e., the last feature set) are the variables that mostly contribute to the predictive performance. 

To further explore this correlation, Fig.~\ref{fig:coef_changes} shows the changes of feature coefficient between performance-optimized predictive model (red; $\theta^{lg}$) and fair model (green; $\theta^{fair}$). We notice that both temperature and sofa have the most impact on EOD difference but have very little coefficient changes. Race is the sensitive attribute; however, the coefficient for race is only changed to a small extent and is very close to 0. Moreover, fair model stresses more weights on those least influential variables (e.g., aspiii, charlson comorbidity index, and age), which also reflects why those least influential variables are the most contributed predictive variables identified using SHAP (two lower subplots in Fig.~\ref{fig:shap-rankings}). The reason for this may be because we add a performance regularization when optimizing the performance-optimized predictive model into fair model; under this given condition, altering the coefficient on important predictive variables will decrease the performance, so in order to reduce the EOD the other variables are the choices during the optimization. This observation aligns with our aforementioned observations, where features with less influence on prediction performance contribute more to improving fairness.

\section{Conclusion and Future Work}

In this study, we employ the transfer learning technique to optimize a performance-optimized model for the prediction of sepsis mortality, aiming to achieve fairness in the model's predictions. Crucially, we introduce a novel permutation-based algorithm designed to elucidate the role of each individual feature in contributing to the fairness of the model. Through a detailed analysis of feature importance and the differential impact of feature permutation on the model's fairness, this approach enables us to effectively bridge the observed gap between model interpretability and fairness. Our findings reveal an under-investigated yet significant relationship between feature importance and the enhancement of model fairness: the more a feature contributes to the predictive performance, the less the feature contributes to improving fairness.

\subsubsection{Acknowledgement.}  This work was supported in part by the National Science Foundation under the Grants IIS-1741306 and IIS-2235548, and by the Department of Defense under the Grant DoD W91XWH-05-1-023.  This material is based upon work supported by (while serving at) the National Science Foundation.  Any opinions, findings, and conclusions or recommendations expressed in this material are those of the author(s) and do not necessarily reflect the views of the National Science Foundation.
%
%
\bibliographystyle{splncs04}
\bibliography{main}

\end{document}